\pdfoutput=1

\documentclass[11pt]{article}

\usepackage{EMNLP2022}

\usepackage{times}
\usepackage{latexsym}

\usepackage[T1]{fontenc}

\usepackage[utf8]{inputenc}

\usepackage{microtype}

\usepackage{inconsolata}

\usepackage{graphicx}
\usepackage{amsmath}
\usepackage{amssymb}
\usepackage{booktabs}
\usepackage{multirow}
\usepackage{tabularx}
\usepackage[skip=5pt]{caption}

\usepackage{pifont}
\usepackage{makecell}

\usepackage{algorithmic}
\usepackage[]{algorithm2e}
\RestyleAlgo{ruled} 

\definecolor{Mahogany}{rgb}{0.75, 0.25, 0.0}

\newcommand{\gs}[1]{\textbf{\textcolor{blue}{#1-GS}}}
\newcommand{\kl}[1]{\textbf{\textcolor{red}{#1-KL}}}
\newcommand{\bs}[1]{\textbf{\textcolor{magenta}{#1-BS}}}

\renewcommand{\gs}[1]{}
\renewcommand{\kl}[1]{}
\renewcommand{\bs}[1]{}

\newcommand{\gsedit}[1]{\textcolor{green!90!black}{#1}}
\newcommand{\kledit}[1]{\textcolor{cyan}{#1}}
\newcommand{\bsedit}[1]{\textcolor{Mahogany}{#1}}

\renewcommand{\gsedit}[1]{{#1}}
\renewcommand{\kledit}[1]{{#1}}
\renewcommand{\bsedit}[1]{{#1}}

\newcommand\clearrow{\global\let\rowmac\relax}
\clearrow

\def\v#1{\mathbf{#1}}

\newcommand{\cmark}{\ding{51}}%
\newcommand{\xmark}{\ding{55}}%

\makeatletter
\newcommand\footnoteref[1]{\protected@xdef\@thefnmark{\ref{#1}}\@footnotemark}
\makeatother

\title{Open-Domain Sign Language Translation Learned from Online Video}

\author{Bowen Shi \\
  TTI-Chicago \\
  \texttt{bshi@ttic.edu} \\\And
  Diane Brentari \\
  Univeristy of Chicago \\
  \texttt{dbrentari@uchicago.edu}\\
  \AND
  Greg Shakhnarovich \\
  TTI-Chicago \\
  \texttt{greg@ttic.edu}\\\And  
  Karen Livescu \\
  TTI-Chicago \\
  \texttt{klivescu@ttic.edu}  
  \\}
  
\begin{document}

\maketitle

\begin{abstract}
\kledit{Existing work on sign language translation---that is, translation from sign language videos into sentences in a written language---has focused mainly} on (1) data collected \kledit{in} a controlled environment or (2) data in a specific domain, which 
\kledit{limits the applicability to real-world settings}.
In this paper, we \kledit{introduce OpenASL},  a large-scale American Sign Language (ASL) - English dataset collected from online video sites (e.g., YouTube). \bsedit{OpenASL} contains 288 hours of 
ASL videos in \kledit{multiple domains}
from over 200 signers and is the largest publicly available ASL translation \kledit{dataset} to date. To tackle the challenges \kledit{of} sign language translation \kledit{in realistic settings and without glosses}, 
we propose a \kledit{set} of techniques including sign search as a \kledit{pretext} task for pre-training and fusion of mouthing and handshape features. 
\kledit{The proposed techniques 
produce consistent and large improvements in translation quality, over baseline models based on prior work}.\footnote{Our data and code are publicly available at \url{https://github.com/chevalierNoir/OpenASL}.} 
\end{abstract}

\section{Introduction}
\label{sec:intro}
\begin{figure*}[btp]
    \centering
    \includegraphics[width=\linewidth]{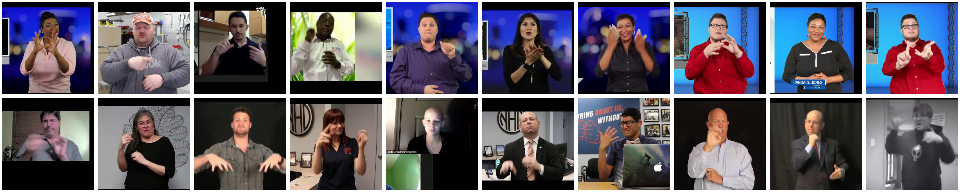}
    \caption{\label{fig:demo-examples}Typical image frames in OpenASL.}
    \vspace{-0.1in}
\end{figure*}

\kl{}
\bsedit{Sign language, a type of visual language that conveys meaning through} \kledit{gestures,} 
is the most widely used 
\kledit{form of linguistic communication among deaf and hard of hearing} people. According to \kledit{the} World Health Organization, over 5\% of the world's population ($\sim$430 million people) suffer from \kledit{disabling} hearing loss.\footnote{\url{https://www.who.int/news-room/fact-sheets/detail/deafness-and-hearing-loss}}
\kledit{Automatic sign language processing}
can facilitate the 
\kledit{daily activities} of deaf people and make \kledit{artificial intelligence}
technologies more accessible to deaf users. For example, \kledit{such techniques would} allow deaf users to interact with \kledit{intelligent virtual assistants} 
using sign language \kledit{and would} support automatic interpretation between sign languages and spoken \kledit{languages.}  \kledit{Interest in sign language research has recently
been growing in the computer vision (CV)~\cite{Bragg2019SignLR,Nikolaos2021comparative,Rastgoo2021sign} \kl{}\bs{} and natural language processing (NLP) communities~\cite{shterionov2021proceedings,yin-etal-2021-including}} 
%

%

%
In this paper, we study sign language translation (SLT),\footnote{Note that "SLT" is often used as an abbreviation for "spoken language technology".  In this paper we use it exclusively for "sign language translation" \kledit{following other recent work.}} the task 
\kledit{of} translating continuous signing video into written language sentences. 
Unlike other sign language processing tasks such as sign spotting~\cite{Buehler2009learning} or continuous sign language recognition \kledit{(sign-to-gloss transcription)}~\cite{Dreuw2007speech}, 
SLT has not been studied until recently~\cite{Camgoz2018neural} and is still restricted to specific domains (e.g., weather \kledit{forecasts}~\cite{Camgoz2018neural}, emergency situations~\cite{ko2019neural}), characterized by small vocabulary size and lack of visual variability.  \kledit{The lack} of large-scale translation datasets in the wild is \kledit{a} central challenge in developing SLT technologies serving real-world \kledit{use} cases.

In terms of translation modeling, \kledit{most} existing SLT approaches~\cite{Camgoz2018neural,camgoz2020multi,camgoz2020multi,Zhou2021ImprovingSL,yin-etal-2021-including,Gan2021SkeletonAwareNS,Chen2022ASM} 
\kledit{rely} on glosses, which \kledit{are a transliteration system for 
sign language}. Annotating sign language videos with glosses is expensive and hard to scale up. Building effective methods for SLT 
\kledit{without glosses is an under-studied challenge.}

In this work, we \kledit{introduce OpenASL,}
a large-scale ASL-English translation dataset. OpenASL has $288$ hours
of real-world ASL videos from over 200 signers, 
\kledit{making it} the largest ASL-English translation dataset to date. 
OpenASL covers \kledit{multiple domains drawn from a mix of news and VLOGs.}
%
%
%
%
\kl{}\bs{} \kl{}\bs{}
To handle challenges in SLT modeling
without glosses, we propose a \bsedit{set} 
of techniques including pre-training with spotted signs and fusion of multiple local visual features, which improves \kledit{over existing} 
SLT baselines by a large margin.

\section{Related Work}
\label{sec:related_work}

\subsection{Datasets for SLT}
There has been a large body of work collecting sign language corpora in general. Here we mainly focus on video-based datasets that can be used for SLT (see \kl{} \bsedit{Table}~\ref{tab:existing-dataset}), which \kledit{contain} paired continuous signing videos and sentences in \kledit{a} written language.

\begin{table}[htp]
\centering
\setlength{\tabcolsep}{2pt}
\resizebox{\linewidth}{!}{
\begin{tabular}{llrrrrl}
\toprule
Dataset & Lang & vocab & \# hours & \# signers & source \\
\midrule 
Phoenix-2014T & DGS & 3K & 11 & 9 & TV \\ 
\cite{Camgoz2018neural}& & &  &  & \\
KETI & KSL & 419 & 28 & 14 & Lab \\
\cite{ko2019neural} & & &  &  & \\
CSL Daily & CSL & 2K & 23 & 10 & Lab  \\ \cite{Zhou2021ImprovingSL}& & &  &  & \\
SWISSTXT-Weather & DSGS & 1K & 1 & - & TV  \\
\cite{camgoz2021content}& & &  &  & \\
SWISSTXT-News & DSGS & 10K & 10 & - & TV  \\
\cite{camgoz2021content}& & &  &  & \\
VRT-News & VGT & 7K & 9 & - & TV  \\
\cite{camgoz2021content} & & &  &  & \\
BOBSL & BSL & 78K & 1467 & 39 & TV  \\ 
\cite{Albanie2021bobsl} & & &  &  & \\
\midrule
  {Purdue RVL-SLLL} & ASL & 104 & - & 14 & Lab  \\
  \cite{Wilbur2006purdue} & & &  &  & \\
{Boston 104} & ASL & 103 & $<1$ & 3 & Lab \\
\cite{Dreuw2007speech} & & &  &  & \\
How2Sign & ASL & 16K & 80 & 11 & Lab  \\ 
\cite{Duarte2021how} & & &  &  & \\
OpenASL & ASL & 33K & 288 & $\sim$220 & Web \\
(Ours) & & &  &  & \\
\bottomrule
\end{tabular}
}
\caption{\label{tab:existing-dataset} Statistics of existing SLT datasets. \bsedit{Example images \kledit{from these} datasets can be found in \kledit{the} Appendix (\kledit{Section}~\ref{sec:app-existing-slt}).} \kledit{The number of signers in OpenASL is approximate, since we cannot determine the identity of the signers for some of the videos.} }
\vspace{-0.05in}
\end{table}

Most of the early datasets~\cite{Wilbur2006purdue,Dreuw2007speech} \kledit{were collected in} a studio-like environment, where 
native signers are 
\kledit{asked to sign some} given content. Recording conditions such as lighting, background and camera perspectives are carefully controlled in such datasets. These corpora provide valuable \kledit{resources,
but} do not account for real-world \kledit{conditions, which has been noted as a limiting factor in recent work on sign language~\cite{yin-etal-2021-including}}.
Moreover, the high cost of data collection also makes studio-based datasets hard to scale up. 

With the advancement of computer vision techniques, there is increasing attention on collecting real-life SLT datasets. Many such datasets~\cite{Camgoz2018neural,camgoz2021content,Albanie2021bobsl} \kledit{are drawn}
from TV programs accompanied \kledit{by} sign language \kledit{interpretation}. Despite being highly realistic compared to studio datasets, they are generally limited to a specific domain. For example, the popular Phoenix-2014T DGS-German benchmark 
\kledit{contains signed} German weather forecasts and 
\kledit{includes only} 11 hours of signing videos from 9 signers. The largest real-world sign language corpus we are aware of is BOBSL~\cite{Albanie2021bobsl}, which consists of 1,467 hours of BBC broadcasts from 39 signers interpreted into British Sign Language (BSL).   However, \kledit{access to the dataset is heavily restricted.} \kl{}\bs{}
%
%

%
\kledit{Unlike} prior datasets, 
\kledit{OpenASL} 
contains a \kledit{mix} of spontaneous and \kledit{(presumably)} interpreted sign language videos. It is collected from online video sites and thus \kledit{contains a diverse set of signers and domains.}
In addition, the annotations we provide are fully accessible to the public.

\subsection{Methods for SLT}

Direct translation from videos of continuous signing is practically appealing and has received 
\kl{} growing interest recently. \kledit{\citet{ko2019neural}} \kledit{study translation of} common Korean sign language sentences (in video) \kledit{that may be} used in an emergency scenario. 
In this specific domain with restricted vocabulary size (419 words), the model can achieve BLEU-4 score higher than 60. 
In a \bsedit{larger-vocabulary} \gs{}\bs{} setting,
\kledit{\citet{Camgoz2018neural}} \kledit{study translation} of German sign language \kledit{weather forecast} videos 
under various labeling setups. In particular, one of \kledit{their} main findings is the drastic improvement achieved \kledit{when} using gloss \kledit{labels} in training an SLT model. It is hypothesized in~\cite{camgoz2020sign} that glosses, as an intermediate representation of sign language, can provide more direct guidance in learning sign language video representation. Therefore, most followup work~\cite{camgoz2020sign,Chen2022ASM,Zhou2021ImprovingSL,Yin2020BetterSL} largely relies on gloss \kledit{sequences} in training.

Given the high cost of gloss labeling, conducting gloss-free SLT is practically appealing \kledit{but} introduces modeling challenges. Glosses, which are monotonically aligned to the video, provide stronger supervision than text in written language \bsedit{translation} %
and facilitate learning of a more effective video representation.  \kledit{On the} Phoenix-2014T benchmark, \kledit{a} model trained without glosses~\cite{Camgoz2018neural} falls behind its counterpart with glosses by over 10.0 (absolute) BLEU-4 score~\cite{camgoz2020sign}. %
Improving translation in real-world sign language video without gloss labels is the modeling focus of this paper. 
\kledit{There is little prior work addressing} SLT without glosses. \bsedit{In a gloss-free setting, }\kledit{\citet{Li2020TSPNetHF} study} the use of segmental structure in translation to boost translation performance. \kledit{\citet{alptekin2020neural} incorporate handshape features} into the translation model.  
\bsedit{In this paper, we consider sign spotting pre-training and fusion of multiple local features for gloss-free translation.}
\kl{}\bs{} \kl{}\bs{} \kl{}\bs{}

A typical SLT model is composed of a visual encoder and a sequence model. The visual encoder maps input video into intermediate visual features. In~\cite{Camgoz2018neural}, a sign recognizer CNN-LSTM-HMM trained with gloss labels was used to extract image features. The continuous sign recognizer was replaced by a CTC-based model in~\cite{camgoz2020sign}. In addition to RGB-based images, pose is also used~\cite{ko2019neural,Gan2021SkeletonAwareNS} as a complementary input modality, which is commonly encoded by graph convolutional neural networks. The sequence models in SLT are usually based on standard sequence-to-sequence models in machine translation with either recurrent neural networks~\cite{Camgoz2018neural} or transformers~\cite{camgoz2020sign,Yin2020BetterSL,Chen2022ASM} as the backbone.

\subsection{Other related work}

\kledit{Two key components of our proposed approach are searching for} spotted signs from video-sentence pairs and fusing multiple local visual features.
There has been \kledit{a substantial} \bsedit{amount of} {prior} 
work~\cite{Buehler2009learning,Albanie2020bsl1k,Varol2021ReadAA,momeni2020watch,shi2022searching} devoted to spotting signs in real-world sign language videos. In contrast to \kledit{this prior work} where sign search is the end goal, here we treat sign spotting as a \kledit{pretext task} in the context of SLT. 

\kledit{The use of} multi-channel visual features has \kledit{also} been previously explored \kledit{for} multiple tasks, including sign spotting~\cite{Albanie2020bsl1k} and continuous sign language recognition~\cite{koller2020weakly}. Specifically for SLT, \kledit{\citet{camgoz2020multi}} \kledit{learn} a multi-channel translation model by including mouthing and handshape features. However, \kledit{these} local modules are trained with in-domain data 
\bsedit{whose labels are inferred from glosses, which makes it  inapplicable for gloss-free translation.}
\gs{}\bs{}  \kledit{In contrast, we utilize models pre-trained on out-of-domain data to extract local features and study the effect of feature transfer to translation.}

\section{The OpenASL Dataset} %
\label{sec:data}

Our videos are collected from 
\kledit{video} websites, mainly YouTube. A large portion of our data consists of ASL news, which come primarily from the %
\kledit{YouTube channels} \texttt{TheDailyMoth} and \texttt{Sign1News}.
We download all videos with English captions in \kledit{these} two channels 
\kledit{through} June 2021. 
The rest \bsedit{of }
\kledit{the dataset is} collected from short \kledit{YouTube videos} 
uploaded 
by the National Association of the Deaf (\texttt{NAD}).
\kl{}
Those videos are mostly in the form of sign VLOGs 
\kledit{of various} types including announcements, daily tips, and short conversations. 

\begin{figure}
\centering
\begin{tabular}{c}
\includegraphics[width=\linewidth]{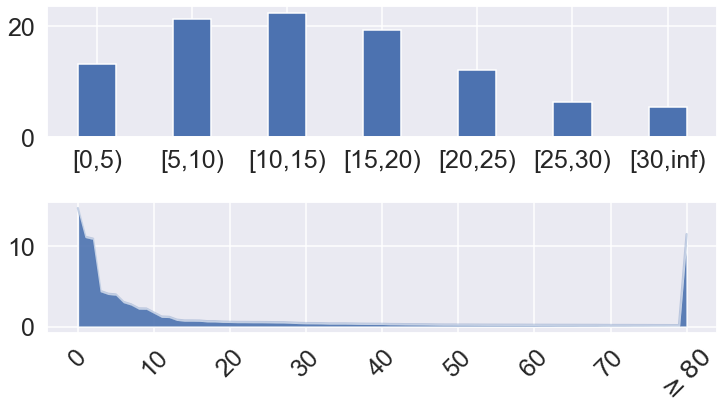} \\
\end{tabular}
\caption{\label{fig:distribution-signer-length} \kledit{Empirical distribution} of 
sentence length 
(upper) and percentage of sequences per signer (lower). The sentence length is in \# words per sentence. \kledit{The signer identity is obtained from metadata of the original video. The sequences where the identity is unknown are not counted.} \kl{}}
\vspace{-0.15in}
\end{figure}

The raw video is divided into \kledit{roughly sentence-sized} clips based on the associated subtitles. Specifically, we split the 
\kledit{transcript} into sentences with the NLTK\footnote{\url{https://www.nltk.org/}}
sentence segmentation tool and retrieve \kledit{the corresponding} (video clip, sentence) pairs. \kledit{This procedure produces}
98,417
\kledit{translation} pairs in \kledit{total, with} 
\bsedit{33,549 unique words.}
\kl{}\bs{} Figure~\ref{fig:distribution-signer-length} shows the distribution of sentence length in our data. 
\kledit{We randomly select 966 and 975 translation pairs from our data} as validation and test sets respectively.\gs{}

\kledit{The annotation of the validation and test sets is manually verified.  Specifically, the English translation and time boundaries of each video clip are proofread and corrected as needed by professional ASL interpreters.}
Each annotator views the video clip and is given the original English sentence from the subtitle for reference. The annotator 
\kledit{marks the corrected} beginning and end of \kledit{the} sentence, and provides \kledit{a corrected} English translation \kledit{if needed} as well as the corresponding gloss sequence. During translation \kledit{of each sentence}, the annotator has access to the whole video in case \kledit{the} context is needed for \kledit{accurate translation}. 

\begin{figure}[htp]
    \vspace{-0.05in}
    \centering
    \includegraphics[width=\linewidth]{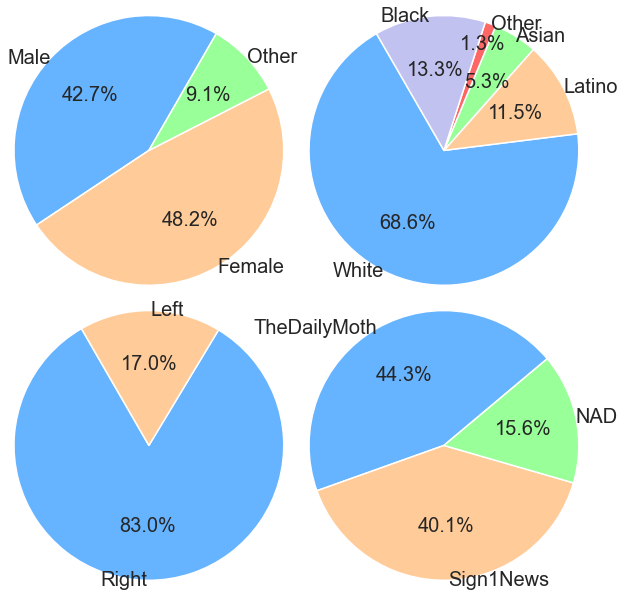}
    \caption{\label{fig:data-distribution}Distribution of \kledit{several}
    properties in \kledit{video clips in a subset of} OpenASL (top left: gender, top right: race, bottom left: handedness, bottom right: \bsedit{sources}).}
    \vspace{-0.05in}
\end{figure}

Figure~\ref{fig:data-distribution} shows the distribution of several properties in our dataset. \kledit{Note that these are not ground-truth labels, but rather approximate labels as perceived by an annotator.  The goal is to give an idea of the degree of diversity in the data, not to provide ground-truth metadata for the dataset.  The label "other" covers a variety of categories, including examples where the annotator is unsure of the label and examples that contain multiple signers.}

\begin{figure}[htp]
    \vspace{-0.05in}
    \centering
    \includegraphics[width=\linewidth]{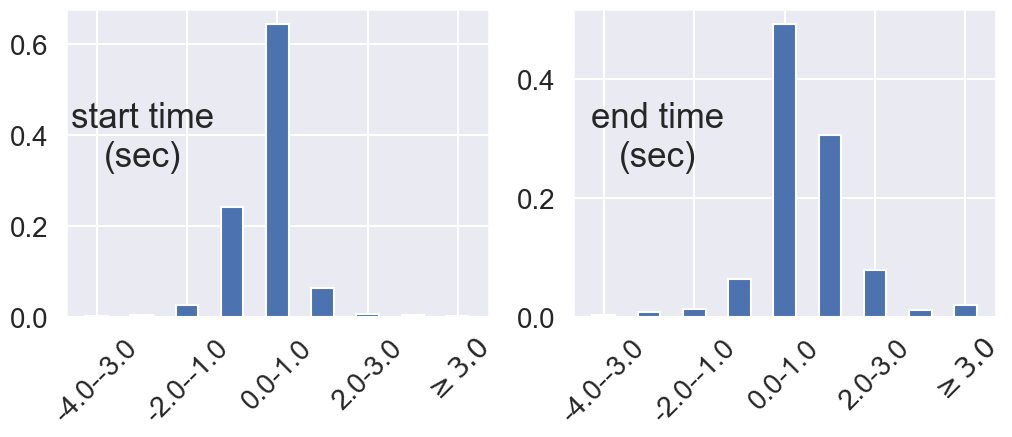}
    \caption{\label{fig:alignment-error}\kledit{Empirical distribution} of alignment \kledit{errors} (in seconds) \kledit{in a manually checked subset of our data} 
    Left: start time, right: end time.}
    \vspace{-0.15in}
\end{figure}

One feature of our data is \kledit{the use} of subtitles associated with the video as the English translation, thus saving effort on human annotation. Subtitled videos have also been employed in prior \kledit{work}~\cite{camgoz2021content,Albanie2021bobsl} for constructing sign language datasets. As prior \kledit{work has mostly focused} on interpreted signing videos where 
content originally in the spoken language is interpreted into sign language, the subtitles \gsedit{used there} are naturally aligned to the audio instead of the signing stream. As is shown in~\cite{bull2021aligning}, there exists a large time boundary shift between the two. The videos used in OpenASL are "self-generated" 
\kledit{rather than} interpreted, \kledit{so} the English subtitles \kledit{are already} 
aligned to the video accurately. As can be seen from \kledit{Figure}~\ref{fig:alignment-error}, the sentence alignment in the subtitles \kledit{is} of overall high quality (\kledit{usually less} than 2 second \kledit{time shifts, although
a small percentage ($<5\%$) are larger)}. \kl{}\bs{}

\bsedit{We measure \kledit{the} degree of agreement between the original and corrected \kledit{translations} using BLEU score~\citep{Papineni2002BleuAM}. The original translation achieves 81.0 BLEU-4 score when it is compared against the corrected one. The high agreement in translation, as well as the small alignment error from Figure~\ref{fig:alignment-error}, shows the overall high quality of the subtitles. Thus to save annotation effort, we do not proofread the training data.}

\section{\kledit{Model and pre-training for gloss-free translation}}
\label{sec:method}
\kl{}

A translation model 
\kledit{maps a sequence of $T$ image frames $\v I_{1:T}$ to a sequence of $n$ words $w_{1:n}$.}
\kledit{
In the most recent state-of-the-art approaches~\citep{camgoz2020sign,Li2020TSPNetHF} for SLT, a visual encoder ${M}_g^v$ first maps
$\v I_{1:T}$ 
to a visual feature sequence $\v f_{1:T^\prime}$,
and a transformer-based sequence-to-sequence model decodes $\v f_{1:T_g}$ into $w_{1:n}$.
Our approach is based on the same overall} \bsedit{ architecture (see Figure~\ref{fig:translation-model}). We further incorporate several techniques for pre-training and local feature modeling, described next.
}
\subsection{Sign spotting pre-training}
\kledit{For} $M^v_g$, we use \kledit{an inflated 3D convnet (I3D) developed for action recognition}~\cite{Carreira2017QuoVA}.
Ideally, the visual encoder 
\kledit{should capture} signing-related visual cues (\kledit{arm movement, handshape, and so on}).  
However, the translated sentence in the target language may not provide \kledit{sufficiently} direct guidance \kledit{for} learning \kledit{the} visual representation\bsedit{, as is observed in prior work~\cite{Camgoz2018neural}.}

To alleviate this issue, we 
\kledit{pre-train} the I3D network on %
\kledit{relevant tasks that provide} more direct supervision \kledit{for} the convolutional layers than \gsedit{full} translation. Specifically,
we pre-train I3D \kledit{for the task of isolated sign recognition on} 
WL-ASL~\cite{Li2020WordlevelDS}, a large-scale isolated \kledit{ASL sign dataset.}
Empirically, we observe considerable gains from isolated sign recognition pre-training (see \kledit{Section}~\ref{sec:app-effect-pretrain-dataset}).

\begin{figure}[htp]
    \centering
    \vspace{-0.1in}
    \includegraphics[width=\linewidth]{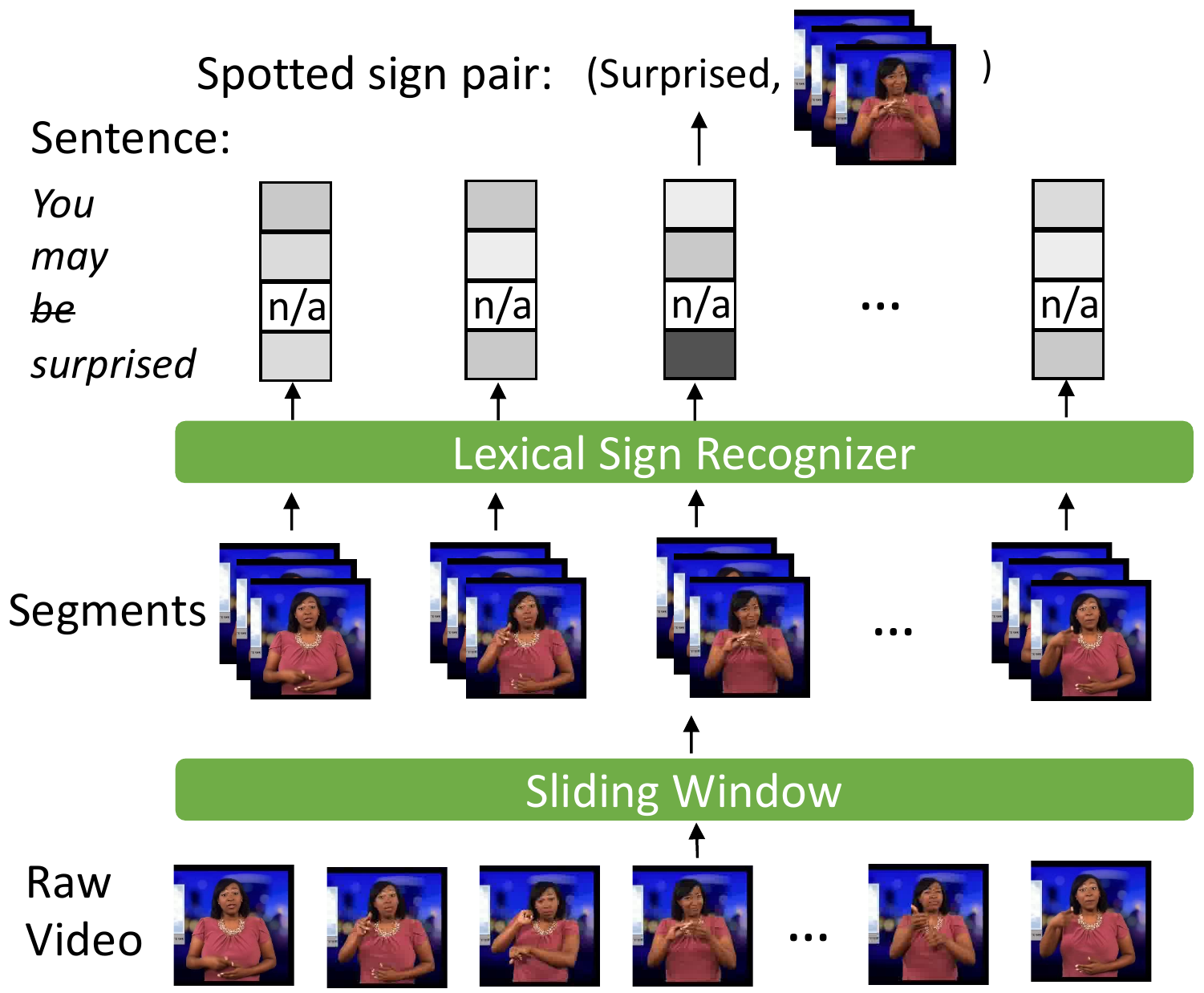}
    \caption{\label{fig:sign-search} Sign spotting. For illustration \kledit{purposes,} only lexical sign search is shown. Fingerspelling sign search works similarly.}
    \vspace{-0.1in}
\end{figure}
Despite the aforementioned benefits, the isolated sign recognition \kledit{pre-training causes} 
two potential problems \kledit{for} the translation model. First, there 
\kledit{is} substantial domain mismatch between isolated signs and \kledit{the continuous} signing data used in translation. The coarticulation in a continuous signing stream is not reflected in isolated sign datasets. 
\kledit{In addition,} the \gsedit{isolated sign} videos are collected from sources such as online \kledit{lexica, which}
usually \kledit{have simpler visual backgrounds 
and less motion blur than}
real-world signing \kledit{video}. 
\kledit{Finally,} existing isolated sign datasets mainly consist of lexical signs and have few instances of fingerspelling. Fingerspelling is used frequently in day-to-day signing and many important content words are commonly fingerspelled. 
Features related to fingerspelling \kledit{may not be encoded well}
due to the lack of fingerspelling-specific pre-training data. 

To mitigate the above issues, we propose to search \kledit{for} 
signs from the signing video (see Figure~\ref{fig:sign-search}). \bsedit{The searched signs} are used to pre-train \kledit{the} visual backbone for translation. \kl{}\bs{} \kl{}\bs{}
The \kledit{search} relies on a lexical sign recognizer ${M}^{l}$ and a fingerspelling recognizer ${M}^{f}$, which \kledit{map} a video segment into \kledit{a} word probability vector $\v p\in[0,1]^V$ ($V$: vocabulary size) and \kledit{a} letter sequence $\v c_{1:|\v c|}$.
Given a translation video-sentence pair ($\v I_{1:T}$, $w_{1:n}$), the 
\kledit{task} 
is to spot 
\textbf{lexical} and \textbf{fingerspelled} signs $\mathcal{P}=\{(\v I_{s_i:t_i}, w_i)\}_{1\leq i\leq |\mathcal{P}|}$\bsedit{, where the $w_i$ are selected from $w_{1:n}$.} \kl{}\bs{}  The search process is described briefly below (see Section~\ref{sec:app-sgin-search} for details). 
%
%

%
%
\kledit{We} generate a list of candidate time intervals for lexical signs and fingerspelling signs respectively with a sliding window approach and a fingerspelling detector $M^d$. For each interval,
we infer its word probability $\v p$ for lexical signs or word hypothesis (i.e., a sequence of characters) $\hat{w}_f$ for fingerspelling. We assign a word from the translated sentence to the target interval if the word probability $p_w$ is high or its edit distance with \kledit{the} fingerspelling hypothesis is low. \kl{}\bs{}

\kledit{Unlike} the isolated sign dataset, the spotted signs are sampled from \bsedit{the same data used for translation \kledit{training}.}
Additionally, the detected fingerspelling signs \kledit{should} also enhance the model's ability
to transcribe signs that are fingerspelled.

\subsection{\bsedit{Hand and mouth} ROI encoding}

In sign language, 
meaning is usually conveyed via a combination of multiple elements \kledit{including motion of the arms, fingers, mouth, and eyebrows}. 
\kledit{The corresponding local regions in the %
image frame play} an important role in %
\kledit{distinguishing} signs. For instance, SENATE and COMMITTEE have the same place of articulation and \kledit{movement;
the difference 
lies only} in the handshape. Furthermore, mouthing (i.e., mouth movement) is commonly used \kledit{for} adjectives or adverbs to add descriptive meaning
~\cite{Nadolske2008Occurence}.  \kl{}
\kledit{Our model's} visual backbone does not explicitly employ 
\kledit{local visual cues.}
In \kledit{principle, learned global features can include sufficient information about the important local cues, but this may require a very large amount of training data.}
\kledit{However, it may be helpful to guide the translation model more explicitly by learning local discriminative features using external tasks.}

Here we focus on learning features for two local visual modalities: handshape and mouthing. To extract handshape \kledit{features}, we train a fingerspelling recognizer\footnote{The implementation is based on~\citep{Shi2019FingerspellingRI} but extended with VGG-19 encoder. See Section~\ref{sec:app-impl-detail} for implementation details.} \kledit{on} two large-scale fingerspelling datasets~\cite{Shi2019FingerspellingRI} and 
\kledit{use} it to extract features for \kledit{the hand region of interest (ROI)}. \kl{}\bs{} \kl{}\bs{}
\kledit{ASL} fingerspelling 
\kledit{includes many handshapres that}
are also used in lexical signs. Recognizing fingerspelling requires distinguishing 
\kledit{quick hand motions} and nuance in finger positions. The features are extracted for both hands in each frame and are concatenated before feeding into the translation model. We denote \kledit{the} hand feature sequence as $\v f_{1:T}^{(h)}$, where $T$ is the video length in frames.

\gsedit{For} mouthing, we 
\kledit{use}
an external English lip-reading model\footnote{We use the publicly available model of~\cite{shi2022avhubert} without any additional training.}~\cite{shi2022avhubert} \kledit{to}
extract 
\kledit{features} $\v f^{(m)}_{1:T}$ \kledit{from the} lip regions of the signer. \kl{}\bs{} \kl{}
 \kledit{Although mouthing in ASL is not used to directly "say" words, we assume there is sufficient shared lip motion between speaking and ASL mouthing.} 
%

\begin{figure}[htp]

    \centering
    \includegraphics[width=\linewidth]{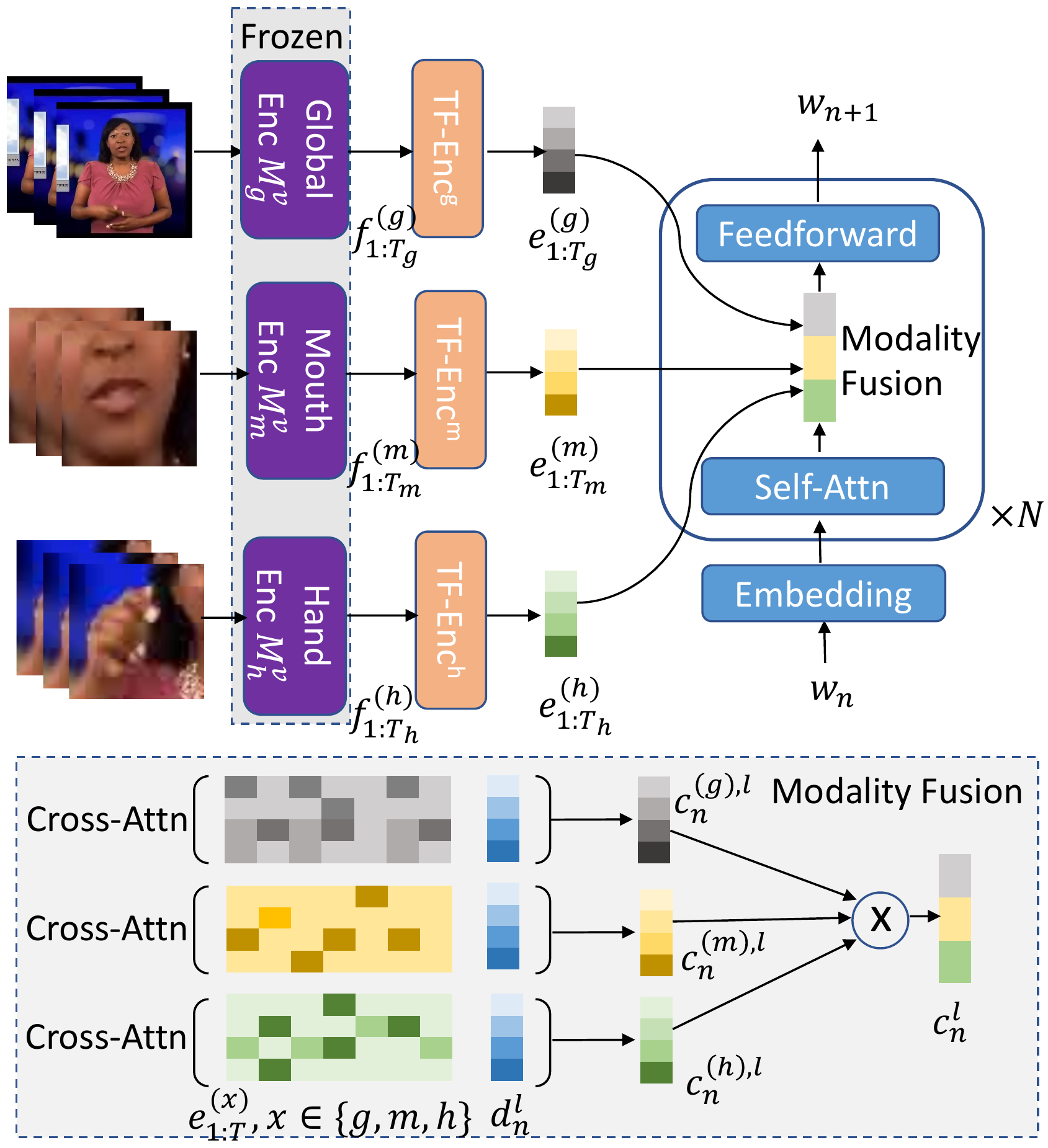}
    \caption{\label{fig:translation-model}Multi-modal sign language translation.}
\end{figure}

\subsection{Fusion and sequence modeling}

\kl{}\bs{}

Given the global/handshape/mouthing feature sequences $\v f^{(g)}_{1:T^\prime}$/$\v f^{(m)}_{1:T}$/$\v f^{(h)}_{1:T}$, the %
\kledit{sequence model 
maps them to} text \kledit{$w_{1:n}$, as
illustrated in Figure~\ref{fig:translation-model}.}
\kledit{Since we have multiple feature sequences each with its own sequential properties},  we adopt three independent transformer~\cite{Vaswani2017AttentionIA} encoders \kledit{for the three types of features}:
\begin{align*}
\label{eq:enc-transformer}
\begin{split}
    \v e^{(x)}_{1:T_x} &= \text{TF-Enc}^{(x)}(\v f^{(x)}_{1:T_x}), x\in\{g,m,h\} \\
\end{split}
\end{align*}
where $\text{TF-Enc}^{(g)}$, $\text{TF-Enc}^{(m)}$, $\text{TF-Enc}^{(h)}$ denote the transformer encoders for global, mouthing and hand feature sequences respectively.  \kl{}
\kledit{For decoding, we use a single transformer decoder that takes all three encoder representations as input.} 
\bsedit{At decoding timestep $n$, we compute modality-specific context vectors:}
\kl{}\bs{} \kl{}\bs{}
\begin{equation*}
\label{eq:cross-attn}
\begin{split}
    \v c^{(x),l}_{n} &= \text{Cross-Attn}^{(x)}(\v d^l_n, \v e^{(x)}_{1:T_x}), x\in\{g,m,h\} \\ 
\end{split}
\end{equation*}
where $\text{Cross-Attn}^{(g)}$, $\text{Cross-Attn}^{(m)}$ and $\text{Cross-Attn}^{(h)}$ are cross-attention layers~\cite{Vaswani2017AttentionIA} for global/mouthing/hand features. 
We concatenate \kledit{the context vectors from the three modalities to form the decoder context vector $\v {c}_n^l=[\v c^{(g),l}_{n},\v c^{(m),l}_{n},\v c^{(h),l}_{n}]$, which is passed to a feedforward layer and then the next layer.  The final layer output is then passed to a linear projection, followed by a final softmax to produce a probability vector over words in the vocabulary.} \kl{}\bs{} \kl{}

%
%
%
%

%
%
%
%

%

%
\section{\kledit{Experiments}}
\label{sec:exp}

\begin{table*}[btp]
\vspace{-0.15in}

\centering
\setlength{\tabcolsep}{3pt}
\resizebox{1\linewidth}{!}{
\begin{tabular}{lrrrrrr|rrrrrr}
\toprule
   & \multicolumn{6}{c|}{DEV} &\multicolumn{6}{c}{TEST} \\ \midrule
Models & {\small ROUGE} & {\small BLEU-1} & {\small BLEU-2} & {\small BLEU-3} & {\small BLEU-4} & {\small BLEURT} & {\small ROUGE} & {\small BLEU-1} & {\small BLEU-2} & {\small BLEU-3} & {\small BLEU-4} & {\small BLEURT} \\
\midrule  
\multirow{2}{*}{\begin{tabular}{@{}l@{}}Conv-GRU \\ {\small \cite{Camgoz2018neural}}$\dagger$\end{tabular}} &
\multirow{2}{*}{16.25} & \multirow{2}{*}{16.72} & \multirow{2}{*}{8.95} & \multirow{2}{*}{6.31} & \multirow{2}{*}{4.82} &
\multirow{2}{*}{25.36} &
\multirow{2}{*}{16.10} & \multirow{2}{*}{16.11} & \multirow{2}{*}{8.85} & \multirow{2}{*}{6.18} & \multirow{2}{*}{4.58} &
\multirow{2}{*}{25.65} \\ 
& & &  &  &  & & &  &  &  \\
I3D-transformer & 18.88	& 18.26 &	10.26 &	7.17 &	5.60 & 29.17 & 18.64 &	18.31 &	10.15 &	7.19 &	5.66 &	28.82  \\ \midrule\midrule
 Ours & \textbf{20.43} &	\textbf{20.10} &	\textbf{11.81} &	\textbf{8.43} &	\textbf{6.57} & \textbf{31.22} &	\textbf{21.02} &	\textbf{20.92} &	\textbf{12.08} &	\textbf{8.59} &	\textbf{6.72}	& \textbf{31.09}  \\ \bottomrule
\end{tabular}
}
\caption{\label{tab:main-result} Translation performance of baseline models and our proposed approach. $\dagger$: based on the public code released by the authors.}
\end{table*}

\subsection{Setup}
\label{sec:exp-setup}

\kl{}\bs{} 
For evaluation, we report BLEU-\{1,2,3,4\}~\cite{Papineni2002BleuAM} and ROUGE~\cite{Lin2004ROUGEAP} \kledit{scores, 
as in prior work on SLT~\cite{Camgoz2018neural,ko2019neural,camgoz2021content}.}
As there is only one English sentence as reference for evaluation, we 
\kledit{also report BLEURT~\cite{sellam2020bleurt} score, a metric that provides a measure of semantic similarity between the prediction and ground truth.}
Implementation details can be found in the appendix (Section~\ref{sec:app-impl-detail}).

\kl{}

\subsection{Main Results}

The performance of our proposed approach is shown in Table~\ref{tab:main-result}.
We compare it to two baseline approaches adapted from prior \kledit{work}. Conv-GRU, which uses ImageNet-pretrained AlexNet as \kledit{a} visual backbone, is an RNN-based sequence-to-sequence model proposed \kledit{by}~\citet{Camgoz2018neural} for sign language translation without glosses. I3D-transformer is a similar model architecture to \kledit{ours, but it uses only}
global visual \kledit{features} and the CNN backbone is pre-trained \kledit{only on the WLASL isolated sign recognition task}. See \kledit{Section}~\ref{sec:app-rwth2014t-perf} in the appendix for the performance of \kledit{these} two baseline methods on the popular DGS-German benchmark Phoenix-2014T.  

\kledit{From the results in Table~\ref{tab:main-result}, we observe:}
(1) Conv-GRU 
\kledit{has} the worst performance among the three models. One key difference 
lies in the data \kledit{used} to pre-train the visual \kledit{encoder:  Conv-GRU}
is pre-trained on ImageNet while the latter two are pre-trained on \kledit{sign language-specific data}. 
\kledit{There are, of course, also differences in the model architecture and training pipeline.  To isolate the effect of sign language-specific pre-training, we 
compare I3D-transformer pre-trained with different types of data, and find that isolated sign pre-training leads to consistent gains.}
See \kledit{Section}~\ref{sec:app-effect-pretrain-dataset} in the Appendix for details.
(2) Our proposed approach achieves the best performance. \kledit{On average,}
the relative gain over I3D transformer is $\sim15\%$ in ROUGE and BLEU scores.
This \kledit{demonstrates}
the effect of including spotted signs in visual backbone pre-training and \kledit{of incorporating the multiple} local visual features.
(3) The performance measured by BLEU, ROUGE and BLEURT scores are consistent \kledit{for} different models.

Despite \kledit{the} improvement over baseline approaches, our \kledit{model's performance is still quite poor.}  
We show some qualitative examples of translated sentence in \kledit{Section}~\ref{sec:app-translation-example} of the Appendix. 
\bsedit{The low performance of current translation models \kledit{has also been} observed in prior work \gsedit{on other sign languages}~\citep{Albanie2021bobsl,camgoz2021content}, highlighting the challenging nature of sign language translation.}  \kl{}

In the next sections,
we analyze the \kledit{effects of the} main components in our model. For \kledit{the purpose of these analyses}, we report BLEU and ROUGE scores on the validation set.

\subsection{Ablation Study}
\label{sec:effect-sign-pretrain}

\textbf{Effect of sign spotting pre-training} In Table~\ref{tab:effect-sign-search}, we compare the performance of models with different pre-training data: WLASL only, WLASL + spotted lexical signs. \kledit{For both models, the visual backbone} is I3D and we do not incorporate local visual features.  
\begin{table}[htp]

\centering
\resizebox{\linewidth}{!}{
\begin{tabular}{lrrrrr}
\toprule
Model & {\small ROUGE} & {\small BLEU-1} & {\small BLEU-2} & {\small BLEU-3} & {\small BLEU-4} \\
\midrule  
iso only & 18.88 & 18.26 & 10.26 & 7.17 & 5.60 \\ 
+spotted & \textbf{19.65} & \textbf{19.72} & \textbf{11.18} & \textbf{8.56} & \textbf{6.51} \\ \bottomrule
\end{tabular}
}
\caption{\label{tab:effect-sign-search} Effect of sign spotting pre-training (iso: isolated sign, spot: spotted signs) on the development set.
}
\end{table}

\kledit{The results show that} sign search 
\kledit{consistently} improves performance. Compared to training with WLASL only, including \kledit{lexical sign and fingerspelling spotting produces} $\sim10\%$ relative \kledit{improvements, averaged across 
metrics.} We attribute \kledit{these gains} to the adaptation of I3D to our translation \kledit{data, which includes coarticulation and visual challenges that the isolated sign data lacks}. \kl{}

\kledit{An alternative strategy could be to fine-tune the visual backbone on our translation data.}
However, 
\kledit{this} strategy downgrades translation performance by a large margin (see \kledit{Section}~\ref{sec:app-finetune-freeze} for details). 
\bsedit{Qualitatively, the spotted sign pairs are high-quality in general (see \kledit{Section}~\ref{sec:search-sign-example}). 
%
}
%
%

%
%

%

%

%

%

\textbf{Effect of local feature incorporation} Table~\ref{tab:effect-local-feature} compares models without local visual \kledit{features}, \kledit{and} with both mouthing and handshape \kledit{features}. All models are pre-trained with spotted signs. Overall the incorporation of local features 
\kledit{produces} 5\% gains in different metrics. \bsedit{The gain is relatively larger in BLEU scores of lower orders (e.g., BLEU-1).} \kl{}\bs{}
See \kledit{Section}~\ref{sec:search-sign-example} for qualitative examples of improved translation \kledit{when} using mouthing \kledit{features}.


\begin{table}[htp]
\centering
\resizebox{\linewidth}{!}{
\begin{tabular}{l|rrrrr}
\toprule
Model & {\small ROUGE} & {\small BLEU-1} & {\small BLEU-2} & {\small BLEU-3} & {\small BLEU-4} \\
\midrule  
global & 19.65 & 19.72 & 11.08 & 8.06 & 6.30 \\ 
+ local & \textbf{20.43} &	\textbf{20.10} &	\textbf{11.81} &	\textbf{8.43} &	\textbf{6.57} \\ \bottomrule
\end{tabular}
}
\caption{\label{tab:effect-local-feature} Effect of incorporating local visual features. \kl{}\bs{.}}

\end{table}

\subsection{Analysis}

For \kledit{a more} detailed analysis \kledit{of our model}, we \kledit{measure its} performance 
on different evaluation subsets, divided by several criteria.

\textbf{Duplicate vs. non-duplicate} Certain sentences appear frequently in our dataset, which leads to duplicated sentences appearing in both training and evaluation. \kledit{The} duplicate sentences account for 10.9\% (105 out of 967) of the dev set and 10.6\% (103 out of 976) of the test set. Most \kledit{of these are sentences that} are used frequently in the \kledit{news,} such as "Hello", "Thank you", \kledit{and "See you tomorrow".}

\begin{figure}[htp]
    \centering
    \includegraphics[width=\linewidth]{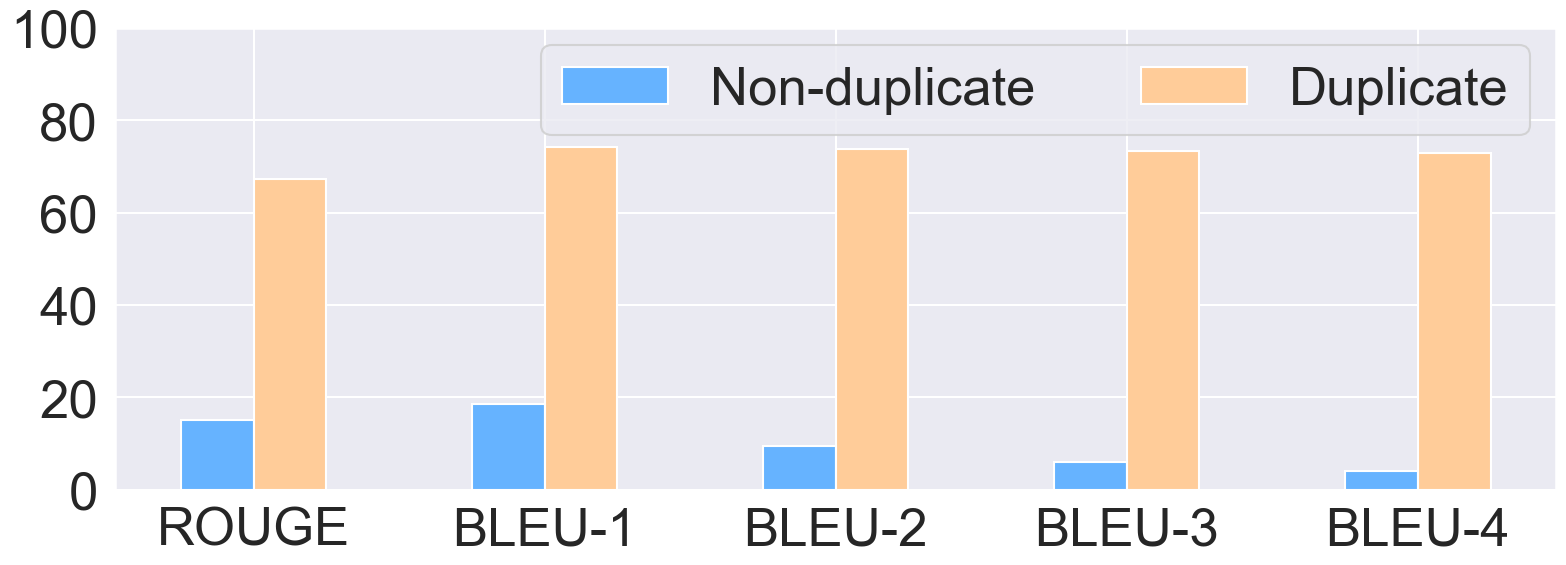}
    \caption{\label{fig:dup-vs-nondup}Comparison of translation performance on duplicate and non-duplicate sentences. Duplicate sentences are \kledit{ones that appear in the training set}.}

\end{figure}

%
\kledit{Our model translates} videos associated with duplicate sentences with high accuracy (see Figure~\ref{fig:dup-vs-nondup}). On \kledit{the duplicate} subset, the BLEU-4 score is \bsedit{72.91}. 
\kledit{Duplicates}
tend to be short clips, which 
\kledit{are} easy for the model to 
memorize.
%
\kledit{In contrast,} the BLEU-4 \kledit{score on the non-duplicate subset} is only 4.09. \kl{}

\begin{figure}[htp]
\vspace{-0.05in}
    \centering
    \includegraphics[width=\linewidth]{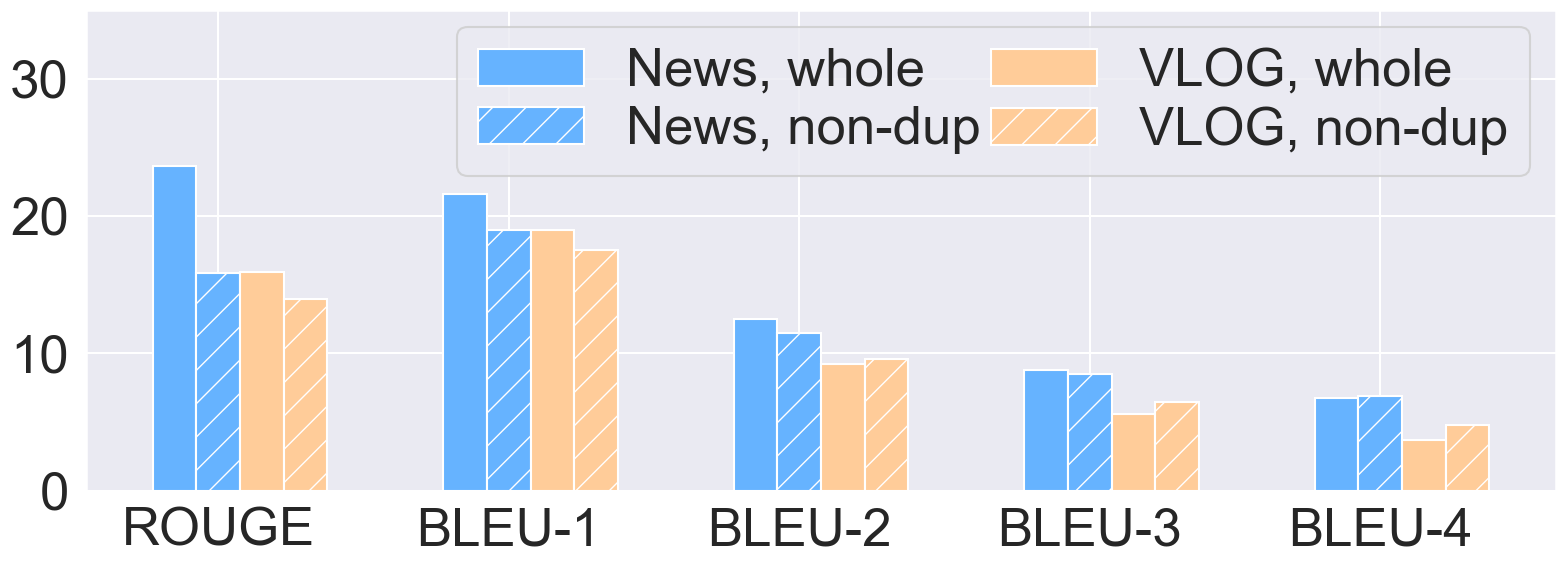}
    \caption{\label{fig:news-vs-vlog}Translation performance for ASL news and VLOGs.}

\end{figure}

\textbf{News vs. VLOGs} Our data are collected from online sign language resources from two categories: news (\texttt{Sign1News} and \texttt{TheDailyMoth}) and VLOGs (\texttt{NAD}). The two sources differ in multiple \kledit{aspects, including} 
visual conditions and linguistic \kledit{content.}
\kledit{In} the dev set, videos from \kledit{these two categories} account for 63.6\%/36.4\% \kledit{of sentences} respectively. We break the performance down according to the source (see Figure~\ref{fig:news-vs-vlog}).  
To avoid the impact of duplicate sentences, we also 
\kledit{perform this comparison 
separately on} non-duplicate sentences. 

Our model 
\kledit{performs better on scripted news} videos regardless of whether the duplicates are included or not, which \kledit{may be} attributed to multiple factors.
On \kledit{the} one hand, \kledit{the} data from 
\kledit{NAD} VLOGs contain a larger set of signers than the news videos. The 
\kledit{variability} in signing among different signers increases the difficulty \kledit{of} translation. \kledit{NAD} VLOG videos also have higher visual variance in terms of image resolution \kledit{and} background diversity. 
\kledit{It is also possible that} the news videos are more likely to be scripted beforehand while the VLOG videos are \kledit{more likely to be} spontaneous. 
Spontaneous ASL videos are \kledit{expected to be} more challenging to translate than scripted videos.

\textbf{Fingerspelling vs. non-fingerspelling} 
In our dev set, $54.7\%$ 
\kledit{of the clips} have at least one fingerspell\kledit{ed word}. 
\kledit{Our model's translation performance on the fingerspelling-free subset is overall higher than on clips with fingerspelling} (BLEU-4: \bsedit{7.74 vs. 6.33}).  \kledit{We expect that} proper nouns, typically fingerspelled in ASL, are difficult to translate \kledit{for} our model. A more detailed analysis can be found in \kledit{Section}~\ref{sec:app-fs-vs-non-fs}. \kl{}\bs{}
%
%
%
%
%
%
\section{Conclusion}
\label{sec:conclusion}
\kl{}

Our work advances sign language translation "in the wild" (i.e., directly translating real-world sign language videos into written language) both (1) by introducing a new large-scale ASL-English translation dataset, OpenASL, and (2) by developing methods for improved translation in the absence of glosses and in the presence of visually challenging data.  OpenASL is the largest publicly available ASL translation dataset to date.  By using online captioned ASL videos, we have been able to collect a large amount of high-quality and well-aligned translation pairs (as verified by professional translators) that represent a wide range of signers, domains, and visual conditions.  Our translation approach, which combines pre-training via sign spotting and multiple types of local features, outperforms 
alternative methods from prior \kledit{work} by a large margin.  Nevertheless, the overall translation quality for sign language videos, in both our work and prior work, is significantly lower than that of machine translation for written languages.  There is therefore much room for future improvement, and we hope that OpenASL will enable additional progress on this task.


\kl{}

\newpage

\section*{Limitations}
\label{sec:limitation}
Despite being the largest ASL translation dataset \kledit{to date}, OpenASL is still of relatively small scale compared to commonly used translation corpora for written languages. Due to resource constraints, we \kledit{provide only} one English translation \kledit{for each} 
video \kledit{for the time being.}
\kledit{Future work may augment the dataset with multiple English translations per video.} \kledit{In addition, although we strive to collect a diverse dataset, we do not have ground-truth labels for signer gender, race, and handedness, so we cannot be certain about the distribution of these properties in OpenASL.}
\bsedit{In terms of methodology, the proposed lexical sign search relies on \kledit{the availability of} isolated sign data. Moreover, \kledit{our approach} may have difficulty in handling ASL signs \kledit{that} do not have an equivalent word in English (e.g., PAH!).}
Finally, the overall quality of English \kledit{translations produced by} our model is still very low, which highlights the \kledit{challenging nature of} open-domain sign language translation.

\section*{Ethics Statement}
\label{sec:ethics-statement}
The copyright \kledit{for} all videos in our dataset \kledit{belongs} to their respective owners. The video URL, timestamps and English translations are released under a Creative Commons BY-NC-ND 4.0 license.
Our data are collected from online video sites, \kledit{and} the signers may not be representative of \kledit{the} general deaf \kledit{or ASL signing} population. Please be \kledit{aware} of unintended racial or gender bias caused by this fact. 
\kledit{Finally,} the translation model proposed in this paper \kledit{still has} low performance and hence \kledit{is} unable to serve as an alternative to human interpreters in real-life scenarios.

\bibliography{anthology,custom}
\bibliographystyle{acl_natbib}

\newpage

\appendix
\section{Appendix}
\label{sec:appendix}

\subsection{Existing datasets}
\label{sec:app-existing-slt}
\begin{figure}[htp]
    \centering \setlength{\tabcolsep}{2pt}
\resizebox{1\linewidth}{!}{
    \begin{tabular}{ccc}
    \toprule
    {\small Phoenix-14T} & {\small KETI} & {\small CSL-Daily} \\
    {\small \citep{Camgoz2018neural}} & {\small \citep{ko2019neural}} & {\small \citep{Zhou2021ImprovingSL}} \\
        \includegraphics[width=0.32\linewidth,height=0.32\linewidth]{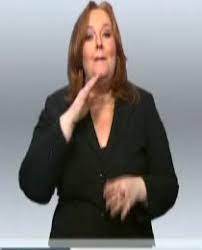} & \includegraphics[width=0.32\linewidth,height=0.32\linewidth]{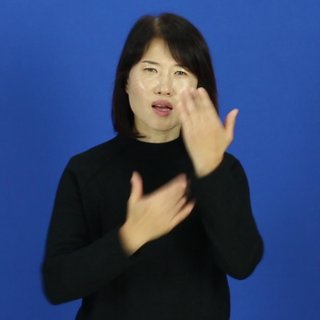} & \includegraphics[width=0.32\linewidth,height=0.32\linewidth]{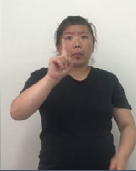} \\
        \midrule
    {\small SWISSTXT} & {\small VRT} & {\small BOBSL} \\
    {\small \citep{camgoz2021content}} & {\small \citep{camgoz2021content}} & {\small \citep{Albanie2021bobsl}} \\
        \includegraphics[width=0.32\linewidth,height=0.32\linewidth]{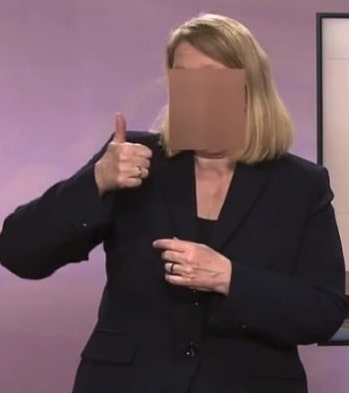} & \includegraphics[width=0.32\linewidth,height=0.32\linewidth]{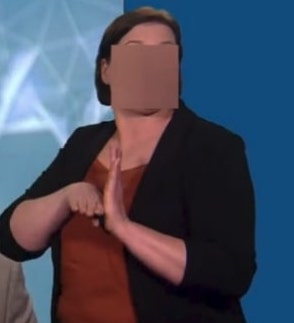} & \includegraphics[width=0.32\linewidth,height=0.32\linewidth]{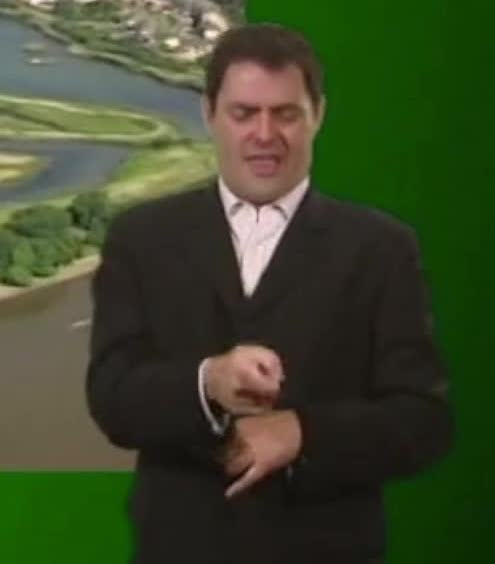} \\
        \midrule
    {\small Purdue-RVL-SLLL} & {\small Boston-104} & {\small How2sign} \\
    {\small \citep{Wilbur2006purdue}} & {\small \citep{Dreuw2007speech}} & {\small \citep{Duarte2021how}} \\
        \includegraphics[width=0.32\linewidth,height=0.32\linewidth]{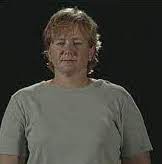} & \includegraphics[width=0.32\linewidth,height=0.32\linewidth]{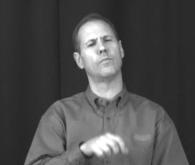} & \includegraphics[width=0.32\linewidth,height=0.32\linewidth]{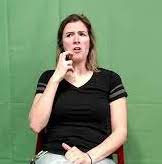} \\
        \bottomrule
    \end{tabular}
    }
    \caption{Image frames from existing SLT datasets.}
    \label{fig:existing-slt-datasets}
\end{figure}

Figure~\ref{fig:existing-slt-datasets} shows typical image frames in existing SLT datasets. Overall, existing SLT data are collected in controlled environments with relatively little visual variability (e.g., background, lighting).

\subsection{Instructions for meta annotation}
\label{sec:annotation-instruction}
For the meta annotation, the annotators were provided the complete videos, and were required to annotate the following information: (1) Number of signers appearing at any time in the video (“one” or “multi” for multiple signers) (2) Name of signer, if it appears in the video description. If you cannot find it, mark this field “UNK” (3) Handedness (left or right; if multiple signers, mark this field “multi”) (4) Perceived gender (“male”, “female”, or “other/unknown”) (5) Perceived race (white/caucasian (including latino/hispanic), black/African American, south Asian, east Asian, other/unknown) (6) Perceived age group (child, young adult, middle aged, older adult) (7) Perceived ASL proficiency (native or near-native, high proficiency, low proficiency).

\subsection{Sign Search}
\label{sec:app-sgin-search}
The search process for lexical and fingerspelling signs is detailed in Algorithm~\ref{alg:sign-search}.

\begin{algorithm}[!h]
\caption{Sign Search}\label{alg:sign-search}
\SetKwInput{KwData}{Data}
\SetKwInput{KwModel}{Model}
\SetKwInput{KwParam}{Hyperparameters}
\SetKwInput{KwResult}{Output}
\KwData{Translation dataset $\mathcal{D}^t$}
\KwModel{isolated sign recognizer $M^{l}$, fingerspelling recognizer $M^f$, fingerspelling detector $M^d$}
\KwParam{lexical/fingerspelling threshold $\delta_l$/$\delta_f$}
\KwResult{Spotted lexical and fingerspelling sign dataset $\mathcal{D}^s$}
\SetKwFunction{FMain}{SearchSign}
\SetKwProg{Fn}{Function}{:}{}
\Fn{\FMain{$\mathcal{D}^t$, $M^{\{l,d,f\}}$, $\delta_{\{l,f\}}$}}{
$\mathcal{D}^s\gets \emptyset$\;
\For{$(\v I_{1:T}, w_{1:L})\in\mathcal{D}^t$}
{
Sliding windows $\Omega_s=\{(s_i, e_i)\}_{1:|\Omega_s|}$\;
\For{$(s, e)\in \Omega_s$}{
Let $\v p\gets M^l(\v I_{s:e})$ be probability vector\;
Let $\tilde{w}_{1:L^\prime}\gets$ the subset of $w_{1:L}$ within the vocabulary of $M^l$\;
Let $\v q\gets (p_{\tilde{w}_1}, p_{\tilde{w}_2},...,p_{\tilde{w}_{L^\prime}})$\;
Let $k\gets \text{argmax}\{\v q\}$\;
\If{$q_k>\delta_l$}{
$\mathcal{D}^s\gets \mathcal{D}^s\cup\{(\v I_{s:e}, \tilde{w}_k)\}$
}
}
Fingerspelling proposals $\Omega_f=\{(s_i, e_i)\}_{1:|\Omega_f|}=M^d(\v I_{1:T})$\;
\For{$(s, e)\in \Omega_f$}{
Word hypothesis $\hat{w}_f=M^f(\v I_{s:e})$\;
Accuracies $\v y=(A(\hat{w}_f, w_1),...,A(\hat{w}_f, w_L))$, $A$: letter accuracy function\;
Let $k\gets \text{argmax}\{\v y\}$\;
\If{$y_k>\delta_l$}{
$\mathcal{D}^s\gets \mathcal{D}^s\cup\{(\v I_{s:e}, w_k)\}$
}
}

}
\KwRet $\mathcal{D}^s$\;
}
\end{algorithm}

\subsection{Implementation details}
\label{sec:app-impl-detail}

\textbf{Preprocessing}  For training, we use the time boundaries in the associated video caption to segment raw videos into short clips. 
We extend the time boundaries of \kledit{each} video clip by 0.5 second \kledit{at} both the beginning and \kledit{the} end to reduce the proportion of potential missing frames caused by misalignment between subtitle and signing video. 
Each video clip is cropped to include only the signing region of the target signer. Specifically, we employ \kledit{the} DLIB face detector~\cite{King2009DlibmlAM} to detect the face of the target signer and crop an ROI centered on the face which is 4 times \kledit{the size} of the original bounding box. In case \kledit{there are multiple faces} detected, we employ a simple heuristic to determine the target face track (tracks with the highest optical flow~\cite{Farnebck2003TwoFrameME} magnitude). The selected ROI is resized to $224\times 224$.
\bsedit{We use words as output units and \kledit{keep} words appearing at least twice in the training set in the vocabulary (21,475 words). } \kl{this seems to conflict with the 30K mentioned earlier}

\textbf{Visual Backbone} For global visual feature extraction, we adopt I3D network~\cite{Carreira2017QuoVA} as our backbone. 
The I3D, pre-trained on Kinetics-400~\cite{Carreira2017QuoVA} is further fine-tuned on WLASL~\cite{Li2020WordlevelDS}, an isolated ASL sign dataset with 14,289 isolated training videos \kledit{of} 2000 distinct ASL signs.
The isolated sign recognizer achieves 42.6\% accuracy on \kledit{the} WLASL test set.

For hand feature extraction, we train a fingerspelling recognizer on \kledit{the} ChicagoFSWild~\cite{Shi2018AmericanSL} and ChicagoFSWild+~\cite{Shi2019FingerspellingRI} datasets, which include 61,536 ASL fingerspelling sequences.
The recognizer is based on \kledit{a} Conv-LSTM architecture~\cite{Shi2021FingerspellingDI}
 consisting of \kledit{the} first 11 conv layers of VGG-19 followed by a one-layer Bi-LSTM with 512 hidden units per direction.  The model is trained with CTC loss~\cite{Graves2006ConnectionistTC} and achieves 64.5\% letter accuracy on \kledit{the} ChicagoFSWild test set. In order to extract hand features on our data, we use \kledit{the} HR-Net whole-body pose estimator~\cite{Sun2019DeepHR} to detect \kledit{the} hands of the signer and extract features in the hand ROI. Features for left and right \kledit{hands} are concatenated before feeding into the translation model. 

To obtain mouthing feature, we employ AV-HuBERT~\cite{shi2022avhubert}, a state-of-the-art lip reading model for English. The mouth ROI, cropped and resized to $96\times 96$ based on the facial landmarks detected with DLIB facial keypoint detector~\cite{King2009DlibmlAM}, are fed into the lip reading model for feature extraction.

\textbf{Sign Search} To search lexical signs, we run inference with the aforementioned I3D isolated sign recognizer on 32-frame windows. The window is swept across the whole video clip at a stride of 8 frames. To search fingerspelling, we use the off-the-shelf fingerspelling detector~\cite{Shi2021FingerspellingDI} trained on raw ASL videos of ChicagoFSWild+, which has achieved 0.448 AP@0.5. The aforementioned fingerspelling recognizer is used for searching fingerspelling signs. We keep proposals with confidence score higher than 0.5. The thresholds $\delta_l$/$\delta_f$ are tuned to be 0.6/0.2 respectively. The total number of signs detected from our translation data is 32,602. 
We combine WLASL and the spotted signs for pre-training I3D (see section~\ref{sec:effect-sign-pretrain}). 
The model is trained with SGD for 50 epochs at batch size of 8. The learning rate and momentum of SGD  are 0.01 and 0.9 respectively. The learning rate is reduced to half at epoch 20 and 40. 

\textbf{Sequence Model} The visual backbones are frozen in training translation model. Both transformer encoder and decoder have 2 layers with 512 hidden dimension and 2048 feedforward dimension. The model is trained with Adam~\cite{Kingma2015AdamAM} for 14K iterations at batch size of 64. The learning rate is linearly increased to 0.001 for the first 2K iterations and then decayed to 0 in the later iterations. At test time, we use beam search for decoding. The beam width and length penalty are tuned on the validation set. 

\textbf{Real-time performance} \kledit{Although real-time performance is not a goal of this work, we note that the} whole proposed system (including all pre-processing such as mouth/hand ROI estimation) processes $\sim$25 frames per second on average for a sign language video from scratch on one RTX A4000 GPU. 

\subsection{Baseline performance on Phoenix-14T}
\label{sec:app-rwth2014t-perf}

Table~\ref{tab:ph14t-perf} shows the performance of the two baseline approaches on Phoenix-14T. In contrast to results on OpenASL, I3D-transformer does not generally outperform Conv-GRU, which is probably due to the linguistic discrepancy between the isolated sign data used to pre-train I3D (WLASL: ASL) and the translation data (Phoenix-14T: DGS).

\begin{table}[htp]
\centering
\setlength{\tabcolsep}{2.5pt}
\resizebox{\linewidth}{!}{
\begin{tabular}{c|rrrrr}

\toprule
Model & ROUGE & BLEU-1 & BLEU-2 & BLEU-3 & BLEU-4 \\
\midrule  
Conv-GRU & \textbf{31.80} & \textbf{32.24} & \textbf{19.03} & 12.83 & 9.58 \\ 
\citep{Camgoz2018neural} &  &  &  &  & \\
I3D-transformer & 27.92	& 26.88	& 18.18 &	\textbf{13.42} &	\textbf{10.66} \\ \bottomrule
\end{tabular}
}
\caption{\label{tab:ph14t-perf} Performance of baseline approaches on Phoenix-14T test set.}
\end{table}

\subsection{\kledit{Does fine-tuning the visual encoder help?}}
\label{sec:app-finetune-freeze}

By default, the visual backbone is frozen in training the translation model. Table~\ref{tab:finetune-vs-freeze} compares pre-training and fine-tuning I3D visual encoder for translation. 
Fine-tuning visual backbone deteriorates the model performance by a large margin. This probably suggests that the proposed benchmark is in a low-resource regime, which does not have enough data for full fine-tuning. We hypothesize that the paired text does not provide strong supervision to learn visual encoder, thus leading to performance degradation. Training a fully end-to-end SLT model potentially requires much larger amount of paired data.

\begin{table}[htp]
\centering
\setlength{\tabcolsep}{2pt}
\resizebox{\linewidth}{!}{
\begin{tabular}{c|rrrrr}
\toprule
Fine-tuning? & ROUGE & BLEU-1 & BLEU-2 & BLEU-3 & BLEU-4 \\
\midrule  
\xmark & \textbf{18.88} & \textbf{18.26} & \textbf{10.26} & \textbf{7.17} & \textbf{5.60} \\ 
\cmark & 18.91 & 16.95 & 9.12 & 5.87 & 4.38 \\ \bottomrule
\end{tabular}
}
\caption{\label{tab:finetune-vs-freeze} Comparison between fine-tuning and freezing visual backbone.}
\end{table}

\subsection{Which pre-training data to use?}
\label{sec:app-effect-pretrain-dataset}

To show the effect of isolated sign pre-training, we compare I3D pre-trained with Kinetics-400 and WLASL in Table~\ref{tab:pretrain-data}. Kinetics-400~\cite{Carreira2017QuoVA} is a large-scale action recognition dataset including over 306,245 video clips from 400 action categories, while WLASL contains 14,289 clips from 2,000 ASL signs.
Though the size of WLASL is one order of magnitude smaller, using WLASL for pre-training 
outperforms pre-training with Kinetics only \kledit{by a large margin}. Utilizing isolated sign data, despite its amount being scarce, greatly boosts the visual representation that further benefits translation.

\begin{table}[htp]
\centering
\resizebox{\linewidth}{!}{
\begin{tabular}{l|rrrrr}

\toprule
Data & ROUGE & BLEU-1 & BLEU-2 & BLEU-3 & BLEU-4 \\
\midrule  
 K & 13.63 & 12.25 & 5.07 & 3.14 & 2.32  \\
 K$\rightarrow$W & \textbf{18.88} & \textbf{18.26} & \textbf{10.26} & \textbf{7.17} & \textbf{5.60} \\  \bottomrule
\end{tabular}
}
\caption{\label{tab:pretrain-data} Effect of pre-training data (K: Kinetics-400~\cite{Carreira2017QuoVA}),W: WLASL~\cite{Li2020WordlevelDS}).}
\end{table}

\subsection{Fingerspelling vs. non-fingerspelling} 
\label{sec:app-fs-vs-non-fs}

\begin{figure}[htp]
\vspace{-0.15in}
    \centering
    \includegraphics[width=\linewidth]{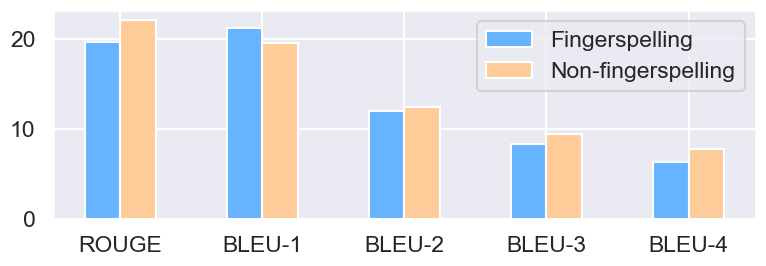}
    \caption{\label{fig:fs-vs-nonfs}Comparison of translation performance between subsets with and without fingerspelling.}
\end{figure}

Fingerspelling is an important component in real-world ASL videos. 
We measure the performance on the subsets with and without fingerspelling respectively. According to Figure~\ref{fig:fs-vs-nonfs}, the translation quality in non-fingerspelling subsets is consistently higher than the other part. Typical fingerspelled words which our model fails to translate are either proper nouns with low frequency in training (e.g., \uppercase{schmidt}, \uppercase{whaley}), or long words (e.g., \uppercase{massachusetts}, \uppercase{salt lake city}).
Though the visual backbone of our translation model is pre-trained with fingerspelling sequences, transcribing the fingerspelling segment(s) is still problematic. As our model is based on whole word, it is incapable of translating words unseen during training. Thus proper nouns, typically fingerspelled in ASL, are difficult to translate by our model.
In practice, we observed many fingerspelled words are simply replaced with $<$UNK$>$. How to improve translation for ASL videos with fingerspelling requires more research.

\vspace{-.1in}
\subsection{Translation examples}
\label{sec:app-translation-example}

We randomly select 15 examples from dev set and compare the model prediction against the reference (see Table~\ref{tab:translation-examples}). The exactly correct translations are mostly short and commonly used sentences in daily communication (e.g., thank you). For longer and more complex sentences, the model frequently fails to capture their general meaning though some keywords can be predicted correctly.

\subsection{Visualization}
\label{sec:search-sign-example}
%
%

%
%

%
%

%
%

\begin{figure*}[htp]
\centering

\resizebox{\linewidth}{!}{
    \begin{tabular}{c}

\hline
husband \\ 
 \includegraphics[width=\linewidth]{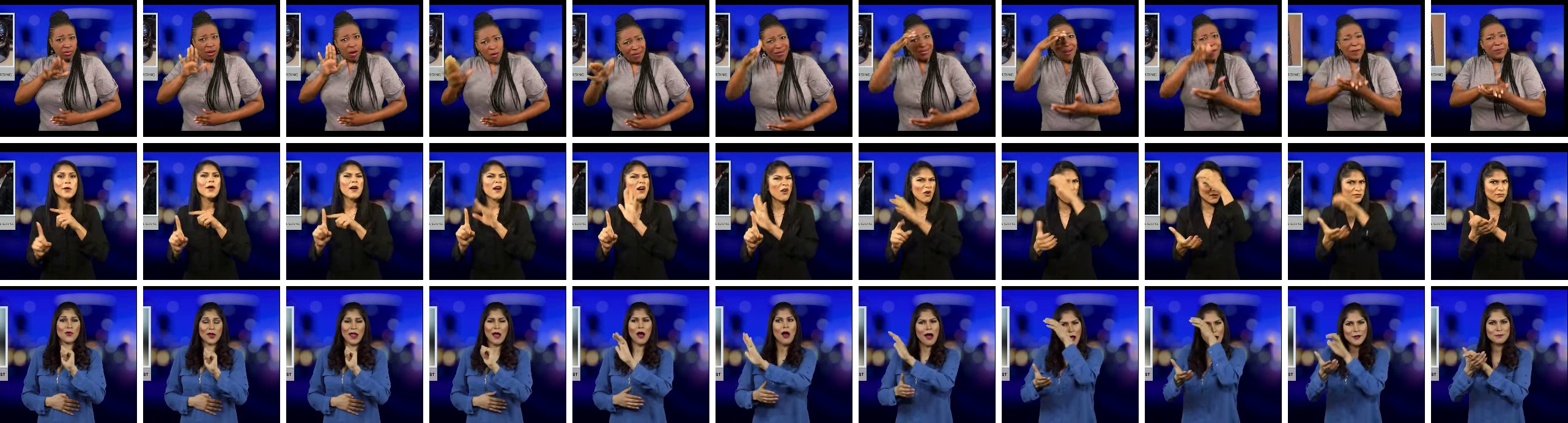}\\ \hline
\hline
before \\ 
 \includegraphics[width=\linewidth]{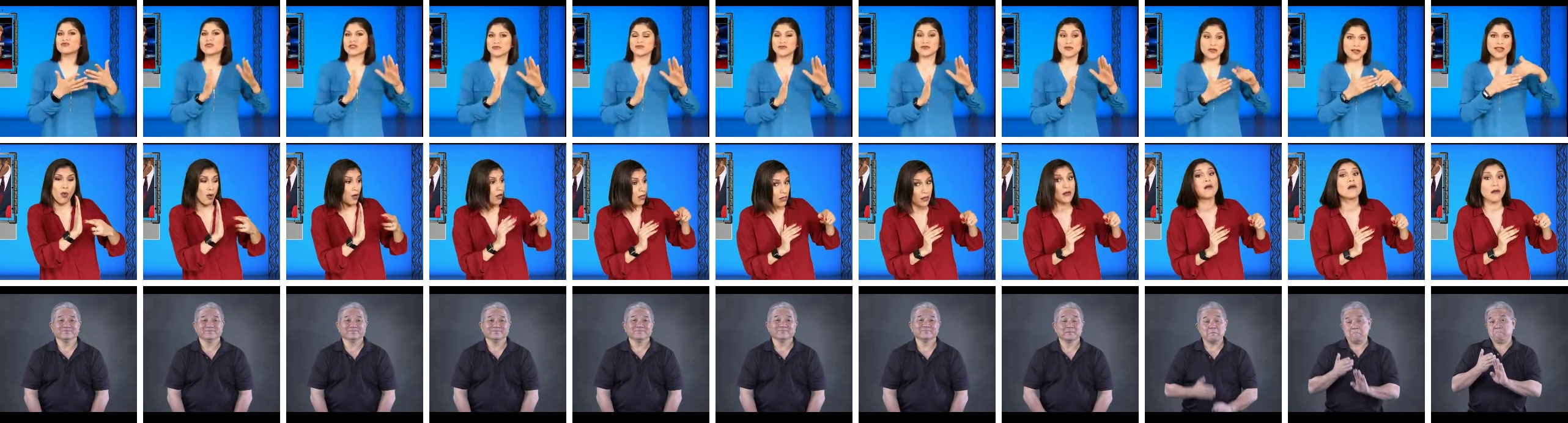}\\ \hline
\hline
university \\ 
 \includegraphics[width=\linewidth]{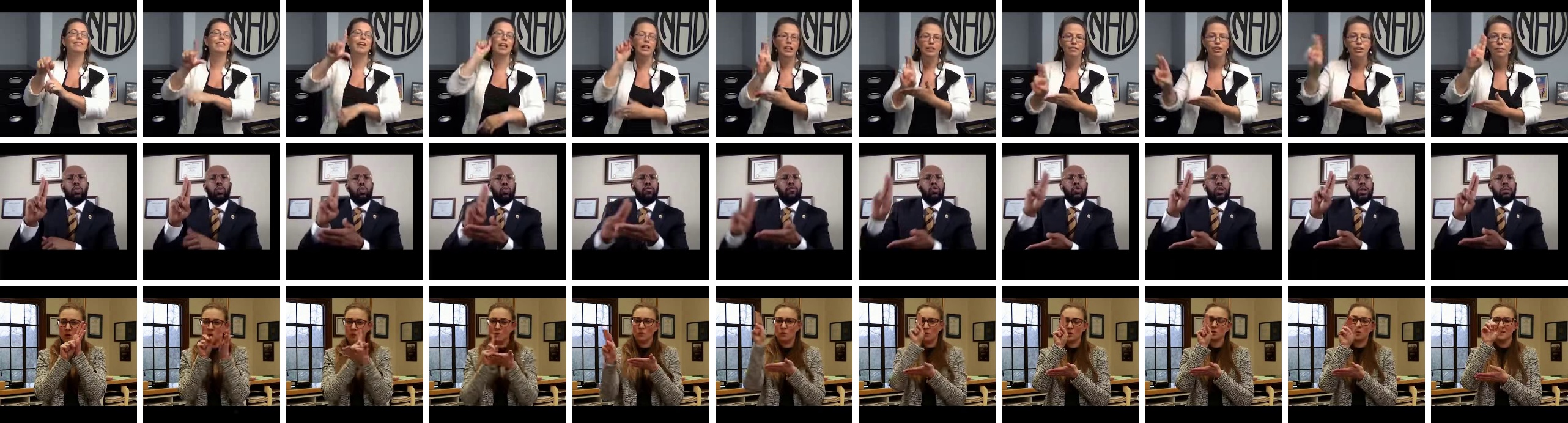}\\ \hline
\hline
FEMA \\ 
 \includegraphics[width=\linewidth]{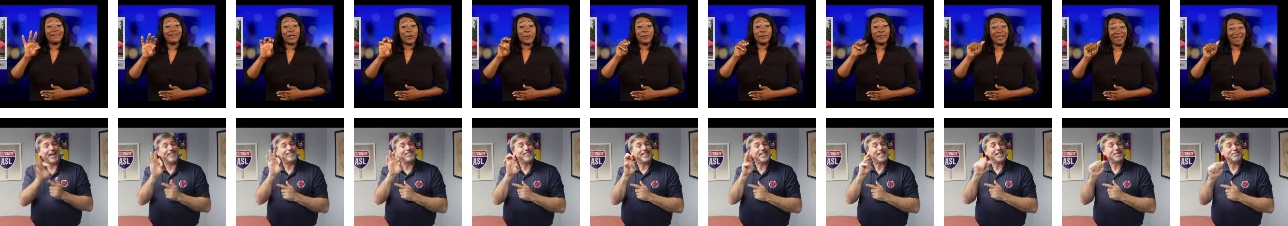}\\ \hline
\hline
WFD \\ 
 \includegraphics[width=\linewidth]{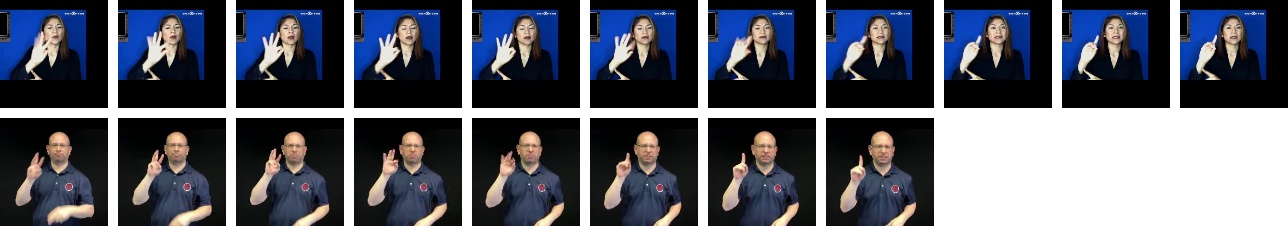}\\ \hline
\hline
BAY \\ 
 \includegraphics[width=\linewidth]{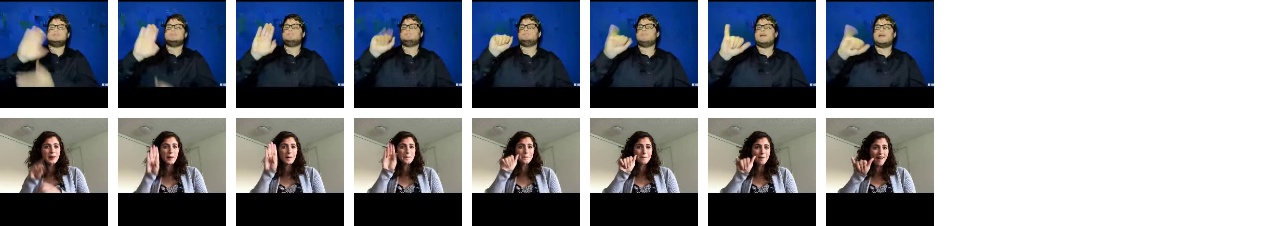}\\ \hline
 
    \end{tabular}
}
\caption{\label{fig:lexical-sign-examples}Qualitative examples of signs spotted by our model (FEMA, WFD and BAY are fingerspelled). }
\end{figure*}

The spotted lexical signs and fingerspelling sequences
are shown in figure~\ref{fig:lexical-sign-examples}. Note that those examples are randomly selected. The spotted signs are mostly accurate. Below are our main observations.
First, in lexical sign spotting, the target clip often includes a partial (or whole) segment from adjacent signs. For instance, the third clip of UNIVERSITY has an extra sign of GALLAUDET. This is due to the fixed window size we use for lexical sign search. 2. False positives occur especially when two signs are of similar appearance. The second clip of BEFORE, which has a similar body posture to BEFORE, is a pointing sign indicating that one thing is happening prior to something else. 3. Using a sophisticated fingerspelling detector enables us to spot fingerspelling sequences more precisely compared to lexical signs. 

\begin{table*}[htp]
\centering
\resizebox{\linewidth}{!}{
    \begin{tabular}{|rl|}
    \hline
\#1 & \\ 
 Ref: & thank you\\ 
 Hyp: & thank you \\ \hline\hline 
\#2 & \\ 
 Ref: & come on\\ 
 Hyp: & come on \\ \hline\hline 
\#3 & \\ 
 Ref: & now i’ve come this far and it ’s a different team\\ 
 Hyp: & how do you feel about it \\ \hline\hline 
\#4 & \\ 
 Ref: & i was there from the beginning to the end and time went by fast\\ 
 Hyp: & the students were thrilled by this \\ \hline\hline 
\#5 & \\ 
 Ref: & i'm here at nad's 50th wow\\ 
 Hyp: & the nad has been <unk> for many years \\ \hline\hline 
\#6 & \\ 
 Ref: & i entered the yap 2018 competition and won\\ 
 Hyp: & the competition was started with ideas \\ \hline\hline 
\#7 & \\ 
 Ref: & you can check out their kickstarter in the link below\\ 
 Hyp: & you can watch the conversation at lake county \\ \hline\hline 
\#8 & \\ 
 Ref: & that is one thing i found interesting and wanted to share with you today\\ 
 Hyp: & i also am the president of the jr. nad conference here \\ \hline\hline 
\#9 & \\ 
 Ref: & those are the different types of bills\\ 
 Hyp: & schools have switched to teaching students \\ \hline\hline 
\#10 & \\ 
 Ref: & dry january has picked up in popularity since it began in 2012\\ 
 Hyp: & krispy kreme is bringing back its original playstation in 2016 \\ \hline\hline 
\#11 & \\ 
 Ref: & we will be happy to respond give you support and listen to your concerns\\ 
 Hyp: & please review and submit your time passion and support this important issue \\ \hline\hline 
\#12 & \\ 
 Ref: & there were videos posted on the internet that showed a person walking on the grass completely engulfed in flames\\ 
 Hyp: & a video shows the officer walking up to his shoulder and before he was shot \\ \hline\hline 
\#13 & \\ 
 Ref: & and people would become carpenters laborers mechanics plowers and farmers\\ 
 Hyp: & the next year 1880 the nad was established in the first operation 30 of the house in 2015 \\ \hline\hline 
\#14 & \\ 
 Ref: & for nad youth programs related information please contact us via facebook at the nad youth programs or email us through\\ 
 Hyp: & you can contact us through our website where you can check our facebook page online at <unk> \\ \hline\hline 
\#15 & \\ 
 Ref: & last week suspects gregory mcmichael and his son travis were arrested and charged with felony murder and aggravated assault\\ 
 Hyp: & last week a black man named <unk> <unk> was arrested and charged with felony murder and aggravated assault \\ \hline\hline 

    \end{tabular}
}
\caption{\label{tab:translation-examples}Qualitative translation examples.  (Ref: reference, Hyp: prediction from our SLT model). Note the examples are randomly chosen without cherrypicking.}
\end{table*}

%

\end{document}